\documentstyle[11pt,fullname]{article}

\newcommand{\rules}{\mbox{$\cal R$}}
\newcommand{\subsumes}{\sqsubseteq}
\newcommand{\subsumed}{\sqsupseteq}
\newcommand{\ns}{\mbox{$I\!\!N$}}
\newcommand{\items}{\mbox{\sc Items}}
\newcommand{\feats}{\mbox{\sc Feats}}
\newcommand{\types}{\mbox{\sc Types}}
\newcommand{\nodes}{\mbox{\sc Nodes}}
\newcommand{\paths}{\mbox{\sc Paths}}
\newcommand{\words}{\mbox{\sc Words}}
\newcommand{\tfss}{\mbox{\sc TFSs}}
\newcommand{\atfss}{\mbox{\sc ATFSs}}
\newcommand{\isdef}{\!\!\downarrow}
\newcommand{\qbar}{\bar{q}}
\newcommand{\Qbar}{\bar{Q}}
\newcommand{\proof}[1]{{\bf Proof:}#1}
\newcommand{\qed}{\hfill $\Box$}
\newcommand{\derives}{\leadsto}
\newcommand{\derivess}{\stackrel{*}{\rightarrow}}
\newcommand{\derivesim}{\rightarrow}

\newtheorem{definition}{Definition}[section]
\newtheorem{theorem}[definition]{Theorem}
\newtheorem{lemma}[definition]{Lemma}
\newenvironment{tfs}[1]{\left[\!\!\!\begin{array}{ll}{\mbox{\bf
#1}}\\}{\end{array}\!\!\!\right]}
\newcommand{\tag}[1]{\fbox{\footnotesize #1}}
\input{psfig}

\title{Off-line Parsability and the Well-foundedness of
  Subsumption\thanks{\ \  To appear in the {\em Journal of Logic,
  Language and Information}}}
\author{Shuly Wintner and Nissim Francez}
\date{Department of Computer Science\\ 
        Technion, Israel Institute of Technology\\ 
        32000 Haifa, Israel}

\begin{document}
\maketitle
\begin{abstract}
Typed feature structures are used extensively for the specification of
linguistic information in many formalisms. The subsumption relation
orders TFSs by their information content. We prove that subsumption of
acyclic TFSs is well-founded, whereas in the presence of cycles
general TFS subsumption is not well-founded. We show an application of
this result for parsing, where the well-foundedness of subsumption is
used to guarantee termination for grammars that are off-line
parsable. We define a new version of off-line parsability that is less
strict than the existing one; thus termination is guaranteed for
parsing with a larger set of grammars.
\end{abstract}
\\
This paper has not been submitted elsewhere in identical or similar form

\section{Introduction}
Feature structures serve as a means for the specification of
linguistic information in current linguistic formalisms such as
LFG~\cite{lfg}, HPSG~\cite{hpsg2} or (some versions of) Categorial
Grammar~\cite{cg}.  This paper focuses on typed feature structures
(TFSs) that are a generalization of their untyped counterparts.  TFSs
are related by {\em subsumption} (see~\cite{carp92}) according to
their information content.  We show that the subsumption relation is
well-founded for acyclic TFSs, but not for cyclic ones. We use this
result to prove that parsing is terminating for grammars that are {\em
off-line parsable}: this proposition is cited, but not proved,
in~\cite{shieber92}. We also suggest a less strict definition for
off-line parsability that guarantees termination in the case of
acyclic TFSs.

This work has originated out of our interest in the theoretical
aspects of parsing with grammars that are based on TFSs
(see~\cite{shuly:tr-lcl-95-1}). While the results presented here are
basically theoretical, we have implemented a system for efficient
processing of such grammars, based on abstract machine techniques;
this work is presented in~\cite{shuly:nlulp-95}. The rest of this
paper is organized as follows: section~\ref{basics} outlines the
theory of TFSs of \cite{carp92}.  In section~\ref{wf} we discuss
the well-foundedness of TFS subsumption. We
sketch a theory of parsing in section~\ref{parsing} and discuss
off-line parsability of TFS-based grammars in
section~\ref{application}.

\section{Theory of feature structures}
\label{basics}
This section summarizes some preliminary notions
along the lines of~\cite{carp92}. While we use the terminology of {\em
typed\/} feature structures, all the results are valid for untyped
structures, that are a special case of TFSs.
\cite{carp92} defines {\em well-typed\/} and {\em totally
well-typed\/} feature structures that are subsets of the set of TFSs;
for generality, we assume nothing about the well-typedness of TFSs below.
However, the larger context of our work is done in a setup where features
are assigned to types through an {\em appropriateness\/}
specification, and hence we retain the term {\em typed\/} feature
structures rather than {\em sorted\/} ones.
For the following discussion we fix non-empty, finite, disjoint sets
\types\ and \feats\ of types and features, respectively, and
an infinite set \nodes\ of nodes, disjoint of \types\ and
\feats, each member of which is assigned a type from \types\
through a fixed typing function $\theta:\nodes\rightarrow\types$.  The
set \nodes\ is `rich' in the sense that for every $t \in \types$, the
set $\{q\in\nodes\mid\theta(q)=t\}$ is infinite.

Below, the meta-variable $T$ ranges over subsets of \types, $t$ --
over types, $f$ -- over features and $q$ -- over nodes. For a partial
function $F$, `$F(x)\isdef$' means that $F$ is defined for the
value $x$. Whenever the result of an application of a partial function
is used as an operand, it is meant that the function is defined for
its arguments. \ns\ denotes the set of natural numbers.

A partial order
$\sqsubseteq$ over $\types$ is a {\bf type hierarchy} 
if it is bounded complete, i.e., if every
up-bounded subset 
$T$ of \types\ has a (unique) least upper bound, $\sqcup T$.
If $t \subsumes t'$, $t$ is said to be more general than $t'$, which
is more specific than $t$. $t'$ is a subtype of $t$.
$\bot=\sqcup\phi$ is the most general type; $\top=\sqcup\types$ is
the most specific, inconsistent type. All occurrences of $\top$ are
identified.

A {\bf feature structure} (over the parameters \nodes, \types\ and
\feats) is a directed, connected, labeled graph consisting of a
finite, nonempty set of nodes $Q\subseteq\nodes$, a root $\qbar\in Q$,
and a partial function $\delta:Q\times\feats\rightarrow Q$ specifying
the arcs, such that every node $q \in Q$ is accessible from $\qbar$.
$A,B$ (with or without subscripts) range over
feature structures and $Q,\qbar,\delta$ (with the same subscripts) over
their constituents.\footnote{Untyped feature structures can be
modeled by TFSs: consider a particular type hierarchy in which the set
of types is the set of atoms, plus the types {\em complex} and
$\bot$, $\bot$ subsumes every other type, and the rest of the types
are incomparable. All features are appropriate for the type {\em
complex\/} only, with $\bot$ as their appropriate value.
Atomic nodes are labeled by an atom, non-atomic
nodes -- by {\em complex} and variables -- by $\bot$.}
Let \tfss\ be the set of all typed feature structures (over the fixed
parameters \nodes, \types\ and \feats).

A {\bf path} is a finite sequence of features, and the set
$\paths=\feats^{*}$ is the collection of paths. $\epsilon$ is the
empty path. 
$\pi,\alpha$ (with or without subscripts) range over paths. The
definition of $\delta$ is extended to paths in the natural way:
$\delta(q,\epsilon)=q$ and $\delta(q,f\pi)=\delta(\delta(q,f),\pi)$.
The paths of a feature structure $A$ are
$\Pi(A)=\{\pi\mid\pi\in\paths$ and $\delta_A(\qbar_A,\pi)\isdef\}$.
Note that for every TFS $A$, $\Pi(A) \neq \phi$ since $\epsilon
\in \Pi(A)$ for every $A$.

A feature structure $A=(Q,\qbar,\delta)$ is {\bf cyclic} if there
exist a non-empty path $\alpha\in\paths$ and a node $q \in Q$ such that
$\delta(q,\alpha)=q$. It is {\bf acyclic} otherwise.
Let \atfss\ be the set of all acyclic TFSs (over the fixed parameters).
%
A feature structure $A=(Q,\qbar,\delta)$ is {\bf reentrant} iff there
exist two different paths $\pi_1,\pi_2\in\Pi(A)$ such that
$\delta(\qbar,\pi_1)=\delta(\qbar,\pi_2)$.

\begin{definition}[Subsumption]
$A_1=(Q_1,\qbar_1,\delta_1)$ {\bf subsumes} $A_2=(Q_2,\qbar_2,\delta_2)$ 
iff there exists a total function $h:Q_1 \rightarrow Q_2$ (a {\bf
subsumption morphism}) such that
\begin{itemize}
\item
$h(\qbar_1)=\qbar_2$
\item
for every $q \in Q_1$, $\theta(q)\sqsubseteq\theta(h(q))$
\item
for every $q \in Q_1$ and for every $f$ such that $\delta_1(q,f)
\isdef$, $h(\delta_1(q,f))=\delta_2(h(q),f)$
\end{itemize}
\end{definition}
The symbol `$\subsumes$' is overloaded to denote subsumption (in
addition to the subtype relation).

The morphism $h$ associates with every node in $Q_1$ a node in $Q_2$
with at least as specific a type; moreover, if an arc labeled $f$
connects $q$ with $q'$ in $A_1$, then such an arc connects $h(q)$ with
$h(q')$ in $A_2$.
Two properties follow directly from the definition: If $A\sqsubseteq
B$ then every path defined in $A$ is defined in $B$, and if two paths
are reentrant in $A$ they are reentrant in $B$.

If two feature structures subsume each other then they have exactly
the same structure. The only thing that distinguishes between them is
the identity of the nodes. This information is usually irrelevant, and
thus an isomorphism is defined over TFSs as follows:
$A$ and $B$ are {\bf alphabetic variants} (denoted $A\sim B$) iff
$A\subsumes B$ and $B \subsumes A$.
$A$ {\bf strictly subsumes} $B$ ($A \sqsubset B$) iff $A \sqsubseteq
B$ and $A \not\sim B$.

If $A$ strictly subsumes $B$ then one of following cases must hold:
either $B$ contains paths that $A$ doesn't; or there is a path in $B$
that ends in a node with a type that is greater than its counterpart
in $A$; or $B$ contains `more reentrancies': paths that lead to the
same node in $B$ lead to different nodes in $A$.
\begin{lemma}
\label{lemma:strict-subsume}
If $A \sqsubset B$ (through the subsumption morphism $h$) then at
least one of the following conditions holds:
\begin{enumerate}
\item
There exists a path $\pi \in \Pi(B) \setminus \Pi(A)$
\item
There exists a node $q \in Q_A$ such that $\theta(q) \sqsubset
\theta(h(q))$,
\item
There exist paths $\pi_1,\pi_2\in\Pi(A)$ such that
$\delta_A(\qbar_A,\pi_1)\neq\delta_A(\qbar_A,\pi_2)$ but
$\delta_B(\qbar_B,\pi_1)=\delta_B(\qbar_B,\pi_2)$. 
\end{enumerate}
\end{lemma}

\section{Well-foundedness}
\label{wf}
In this section we discuss the well-foundedness of TFS subsumption.
A partial order $\succ$ on a set $D$ is {\bf well-founded} iff there
does not exist an infinite decreasing sequence $d_0 \succ d_1 \succ
d_2 \succ \ldots$ of elements of $D$.
We prove that subsumption of acyclic TFSs is well-founded, and
show an example of general (cyclic) TFSs for which subsumption is not
well-founded. While these results are not surprising, and in fact
might be deduced from works such as, e.g.,~\cite{moshier-rounds}
or~\cite{shieber92}, they were not, to the best of our knowledge,
spelled out explicitly before.

\begin{lemma}
\label{lemma:finite}
A TFS $A$ is acyclic iff $\Pi(A)$ is finite.
\end{lemma}
\proof{}
If $A$ is cyclic, there exists a node $q\in Q$ and a non-empty path
$\alpha$ that $\delta(q,\alpha)=q$. Since $q$ is accessible, let $\pi$
be the path from the root to $q$: $\delta(\qbar,\pi)=q$.
The infinite set of paths $\{\pi\alpha^{i}\mid i \ge 0\}$ is contained
in $\Pi(A)$. \\
If $A$ is acyclic then for every non-empty path $\alpha \in \paths$
and every $q \in Q$, $\delta(q,\alpha)\neq q$. $Q$ is finite, and so
is \feats, so the out-degree of every node is finite. Therefore the
number of different paths leaving $\qbar$ is bounded, and hence
$\Pi(A)$ is finite.
\qed

\begin{definition}[Rank]
Let $r:\types\rightarrow\ns$ be a total function such that
$r(t)<r(t')$ if $t \sqsubset t'$. For an acyclic TFS $A$, let
$\Delta(A)=|\Pi(A)|-|Q_A|$ and let $\Theta(A)=\sum_{\pi\in\Pi(A)}
r(\theta(\delta(\qbar,\pi)))$.
Let $rank: \atfss \rightarrow \ns$ be defined by
$rank(A)=|\Pi(A)|+\Theta(A)+\Delta(A)$.
\end{definition} 
By lemma~\ref{lemma:finite}, $rank$ is well defined for acyclic
TFSs. $\Delta(A)$ can be thought of as the `number of
reentrancies' in $A$: every node $q\in Q_A$ contributes
$\mbox{in-degree}(q)-1$ to $\Delta(A)$. For every acyclic TFS $A$,
$\Delta(A)\ge 0$ (clearly $\Theta(A) \ge 0$ and $|\Pi(A)| \ge 0$) and
hence $rank(A)\ge 0$.

\begin{lemma}
\label{lemma:rank}
If $A \sqsubset B$ and both are acyclic then $rank(A)<rank(B)$.
\end{lemma}
\proof{}
Since $A \subsumes B$, 
$\Pi(A) \subseteq \Pi(B)$. Consider the
two possible cases:
\begin{itemize}
\item
If $\Pi(A)=\Pi(B)$, then
\begin{itemize}
\item
$|\Pi(A)|=|\Pi(B)|$;
\item
$\Theta(A) \le \Theta(B)$ by the definitions of $\Theta$ and
subsumption;
\item
$\Delta(A) \le \Delta(B)$ by the definition of subsumption, since
every reentrancy in $A$ is a reentrancy in $B$;
\item
By lemma~\ref{lemma:strict-subsume}, either $\Theta(A)<\Theta(B)$ (if
case (2) holds), or $\Delta(A)<\Delta(B)$ (if case (3) holds).
\end{itemize}
Hence $rank(A)<rank(B)$.
\item
If $\Pi(A) \subset \Pi(B)$ then 
\begin{itemize}
\item
$|\Pi(A)| < |\Pi(B)|$;
\item
$\Theta(A) \le \Theta(B)$ (as above)
\item
it might be the case that
$|Q_A|<|Q_B|$. But for every node $q\in Q_B$ that is
not the image of any node in $Q_A$, there exists a path $\pi$ such
that $\delta_B(\qbar_B,\pi)=q$ and $\pi\not\in\Pi(A)$. Hence
$|\Pi(A)|-|Q_A|\le |\Pi(B)|-|Q_B|$.
\end{itemize}
Hence $rank(A)<rank(B)$.
\qed
\end{itemize}

\begin{theorem}
Subsumption of acyclic TFSs is well-founded.
\end{theorem}
\proof{}
For every acyclic TFS $A$, $rank(A) \in \ns$. By lemma~\ref{lemma:rank},
if $A \sqsubset B$ then $rank(A)<rank(B)$. If an infinite decreasing
sequence of acyclic TFSs existed, $rank$ would have mapped them to an
infinite decreasing sequence in $\ns$, which is a contradiction.
Hence subsumption is well-founded for acyclic TFSs.
\qed

\begin{figure}
\center
\psfig{figure=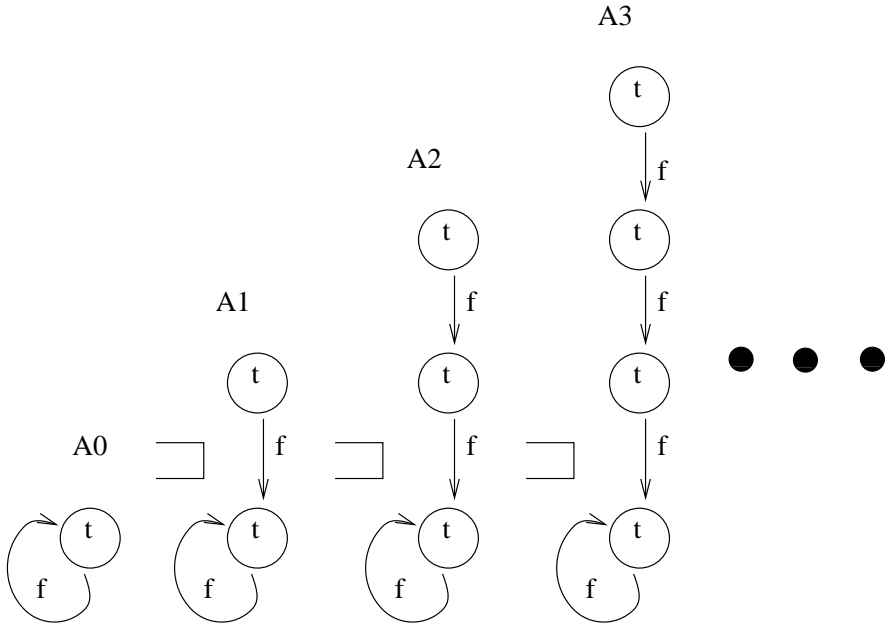}
\caption{An infinite decreasing sequence of TFSs}
\label{fig:sequence}
\end{figure}

\begin{theorem}
Subsumption of TFSs is not well-founded.
\end{theorem}
\proof{}
Consider the infinite sequence of TFSs $A_0,A_1,\ldots$ depicted
graphically in figure~\ref{fig:sequence}.  For every $i \ge 0$, $A_i
\sqsupset A_{i+1}$: to see that consider the morphism $h_i$ that maps
$\qbar_{i+1}$ to $\qbar_{i}$ and $\delta_{i+1}(q,f)$ to
$\delta_i(h(q),f)$ for every $q \in Q_{i+1}$ (i.e., the first $i+1$
nodes of $A_{i+1}$ are mapped to the first $i+1$ nodes of $A_i$, and
the additional node of $A_{i+1}$ is mapped to the last node of
$A_i$). Clearly, for every $i\ge 0$, $h_i$ is a subsumption
morphism. Hence, for every $i\ge 0, A_i \sqsupseteq A_{i+1}$.\\
To show strictness, assume a subsumption morphism $h': Q_i \rightarrow
Q_{i+1}$. By definition, $h'(\qbar_i) = \qbar_{i+1}$. By the third
requirement of subsumption ($h$ commuting with $\delta$), the first
$i+1$ nodes in $A_i$ have to be mapped by $h'$ to the first $i+1$ nodes
in $A_{i+1}$. However, if $q$ is the $i+1$-th node of $A_i$, then
$\delta_i(q)$ leads back to $q$, while $\delta_{i+1}(h'(q))$
leads to the last node of $A_{i+1}$ (the cyclic node), and hence $h'$
does not commute with $\delta$, a contradiction. Hence, $A_i \sqsupset
A_{i+1}$.\\
Thus, there exists a strictly decreasing infinite sequence of cyclic
TFSs and therefore subsumption is not well-founded.
\qed

To conclude this section, note that {\em specification},
which is the inverse relation to subsumption, is not well-founded even
when cyclic feature structures are ruled out. This fact can easily be
seen by considering the sequence of feature structures
$B_0,B_1,\ldots$, where $B_i$ consists of $i+1$ nodes, the first $i$ of
which are labeled $t$ and the last -- $\bot$, and an $f$-arc leads
from every node to its successor. Clearly, $B_i \sqsubset B_{i+1}$ for
every $i>0$, and the sequence is infinite. This is true whether or not
appropriateness constraints are imposed on the feature structures involved.

In the general case, then, given a feature structure $A$ it might be
possible to construct, starting from $A$, both an infinite decreasing
sequence of TFSs (by expanding cycles) and an infinite increasing
sequence (by adding paths).

\section{Parsing}
\label{parsing}
Parsing is the process of determining whether a given string belongs
to the language defined by a given grammar, and assigning a structure
to the permissible strings. A large variety of parsing algorithms
exists for various classes of grammars (for a detailed treatment of
the theory of parsing with grammars that are based on feature
structures, refer to~\cite{shieber92,sikkel,shuly:tr-lcl-95-1}).  We
define below a simple algorithm for grammars that are based on TFSs,
but it must be emphasized that the results presented in this paper are
independent of the particular algorithm; they hold for a wide range of
different algorithms.

To be able to represent complex linguistic information, such as phrase
structure, the notion of feature structures is usually extended.
There are two different approaches for representing phrase structure
in feature structures: by adding special, designated features to the
FSs themselves; or by defining an extended notion of FSs. The first
approach is employed by HPSG: special features, such as DTRS
(daughters), encode trees in TFSs as lists. This makes it impossible to
directly access a particular daughter. \cite{shieber92} uses a variant of
this approach, where a denumerable set of special features, namely
$0,1,\ldots,$ are added to encode the order of daughters in a tree. In
a typed system such as ours, this method would necessitate the
addition of special types as well; in general, no bound can be placed
on the number of features and types necessary to state rules
(see~\cite[p.\ 194]{carp92}).  

We adopt below the other approach: a
new notion of {\em multi-rooted feature structures}, suggested
by~\cite{sikkel}~and~\cite{shuly:tr-lcl-95-1}, is being used.

\begin{definition}[Multi-rooted structures]
A {\bf multi-rooted feature structure} (MRS) is a pair $\langle
\Qbar,G\rangle$ where $G=\langle Q,\delta\rangle$ is a finite, directed, 
labeled graph consisting of a set $Q\subseteq\nodes$ of nodes and a
partial function $\delta:Q\times\feats\rightarrow Q$ specifying the
arcs, and $\Qbar$ is an ordered (repetition-free) set of distinguished
nodes in $Q$ called {\bf roots}.  $G$ is not necessarily connected,
but the union of all the nodes reachable from all the roots in $\Qbar$
is required to yield exactly $Q$.  The {\bf length} of a MRS is the
number of its roots, $|\Qbar|$. $\lambda$ denotes the empty MRS (where
$Q=\phi$ since $|\Qbar|=0$). A MRS is {\bf cyclic} under the same
conditions a TFS is. A MRS is {\bf reentrant} if it contains a node
that can be reached either from two different roots or through two
different paths.
\end{definition}

Meta-variables $\sigma,\rho$ range over MRSs, and $\delta,Q,\Qbar$
over their constituents. If $\sigma=\langle\Qbar,G\rangle$ is a MRS
and $\qbar_i$ is a root in $\Qbar$ then $\qbar_i$ naturally induces a
feature structure $A_i=(Q_i,\qbar_i,\delta_i)$, where $Q_i$ is the set
of nodes reachable from $\qbar_i$ and $\delta_i=\delta|_{Q_i}$. Thus
$\sigma$ can be viewed as an ordered sequence $\langle
A_1,\ldots,A_n\rangle$ of (not necessarily disjoint) feature
structures. We use the two views of MRSs interchangeably.  

Not only
can nodes be shared by more than one element of the sequence; paths
that start in one root can reach a different root.  In particular,
cycles can involve more than one root. Still, it is possible to define
{\em sub-structures\/} of MRSs by considering only the sub-graph that
is accessible from a sub-sequence of the roots.
\begin{definition}[Sub-structure]
The {\bf sub-structure} of $\sigma=\langle
A_1,\ldots,A_n\rangle$, induced by the pair $\langle
i, j \rangle$ and denoted $\sigma^{i\ldots j}$, is $\langle
A_i,\ldots,A_j\rangle$. If $i>j$, $\sigma^{i\ldots j}=\lambda$. If $i=j$,
we use $\sigma^i$ for $\sigma^{i\ldots i}$.
\end{definition}

\begin{definition}[Subsumption of multi-rooted structures]
A MRS $\sigma=\langle \Qbar,G\rangle$ {\bf subsumes} a MRS
$\sigma'=\langle \Qbar',G'\rangle$ (denoted by $\sigma \sqsubseteq \sigma'$) 
if $|\Qbar|=|\Qbar'|$ and there
exists a total function $h:Q \rightarrow Q'$ such that:
\begin{itemize}
\item
for every root $\qbar_i\in\Qbar,h(\qbar_i)=\qbar'_i$
\item
for every $q \in Q$, $\theta(q)\sqsubseteq\theta(h(q))$
\item
for every $q \in Q$ and $f\in\feats$, if $\delta(q,f)\isdef$ then
$h(\delta(q,f))=\delta'(h(q),f)$
\end{itemize}
\end{definition}

Many of the properties of TFSs are easily adaptable to MRSs. Let
$\Pi(\sigma) = \{(\pi,i) \mid \pi \in \paths$, $\qbar$ is the $i$-th root in
$\Qbar_{\sigma}$ and $\delta_{\sigma}(\qbar,\pi)\isdef\}$. Then it is
easy to show that if $\sigma \subsumes \rho$ then $\Pi(\sigma)
\subseteq \Pi(\rho)$ and every reentrancy in $\sigma$ is a reentrancy
in $\rho$. Moreover, if $\sigma \sqsubset \rho$ (strictly) then at
least one of the three conditions listed in
lemma~\ref{lemma:strict-subsume} holds.

The well-foundedness result of the previous section are easily
extended to MRSs as well. Let
$\Theta(\sigma)=\sum_{(\pi,\qbar)\in\Pi(\sigma)}
r(\theta(\delta_{\sigma}(\qbar,\pi)))$ and $\Delta(\sigma) =
|\Pi(\sigma)|-|Q_{\sigma}|$, and the same rank function of TFSs can be
used to show the well-foundedness of (acyclic) MRSs. The reverse
direction is immediate: in the presence of cycles, duplicate the
example of the previous section $k$ times and an infinite decreasing
sequence of MRSs of length $k$ is obtained, for any $k>0$.
For a detailed discussion of the properties of MRSs, refer
to~\cite{shuly:tr-lcl-95-1}. 

Rules and grammars are defined over an additional parameter, a fixed,
finite set \words\ of words (in addition to the parameters \nodes,
\feats\ and \types).  The {\em lexicon\/} associates with every word
$w$ a feature structure $Cat(w)$, its {\bf category}.\footnote{For the
sake of simplicity, we assume that lexical entries are not
ambiguous. In the case of ambiguity, $Cat(w)$ is a set of TFSs. While
the definitions become more cumbersome, all the results still obtain.}
The categories are assumed to be disjoint. The input for the parser,
therefore, is a sequence of (disjoint) TFSs rather than a string of
words.

\begin{definition}[Pre-terminals]
Let $w=w_1\ldots w_n\in\words^{*}$.
$PT_w(j,k)$ is 
defined iff $1\le j,k\le n$, in which case it is the MRS
$\langle Cat(w_j),Cat(w_{j+1}),\ldots,Cat(w_k)\rangle$.
If $j>k$ then $PT_w(j,k)=\lambda$.
\end{definition}
If a single word occurs more than once in the input (that is,
$w_i=w_j$ for $i\neq j$), its category is copied (with remaned nodes)
more than once in $PT$.

\begin{definition}[Grammars] 
A {\bf rule} is an MRS of length greater than or equal to 1 with a designated
(first) element, the {\bf head} of the rule. The rest of the elements
form the rule's {\bf body} (which may be empty).  A {\bf grammar}
$G=(\rules,A_s)$ is a finite set of rules $\rules$ and a {\bf start
symbol} $A_s$ that is a TFS.
\end{definition}
Figure~\ref{fig:grammar} depicts an example grammar (we use AVM notation for
this rule; tags such as $\tag{1}$ denote reentrancy). The type
hierarchy on which the grammar is based is omitted here.

\begin{figure}[hbt]
Initial symbol:
{\scriptsize
\[
\begin{tfs}{phrase}
        CAT: & \begin{tfs}{s} \end{tfs}
\end{tfs}
\]
}
Lexicon:
{\tiny
\[
\begin{array}{ccc}
\mbox{John} & \mbox{her} & \mbox{loves} \\
\begin{tfs}{word}
        CAT:    & \begin{tfs}{n} \end{tfs} \\
        AGR:    & \begin{tfs}{agr}
                        PER:    & \begin{tfs}{3rd} \end{tfs} \\
                        NUM:    & \begin{tfs}{sg} \end{tfs} 
                        \end{tfs} \\
        SEM:    & \begin{tfs}{sem} PRED: & \begin{tfs}{john}\end{tfs} \end{tfs}
\end{tfs}
&
\begin{tfs}{word}
        CAT:    & \begin{tfs}{n} CASE: & \begin{tfs}{acc} \end{tfs}\end{tfs} \\
        AGR:    & \begin{tfs}{agr}
                        PER:    & \begin{tfs}{3rd} \end{tfs} \\
                        NUM:    & \begin{tfs}{sg} \end{tfs} 
                        \end{tfs} \\
        SEM:    & \begin{tfs}{sem} PRED: & \begin{tfs}{she} \end{tfs} \end{tfs}
\end{tfs}
&
\begin{tfs}{word}
        CAT:    & \begin{tfs}{v} \end{tfs} \\
        AGR:    & \begin{tfs}{agr}
                        PER:    & \begin{tfs}{3rd} \end{tfs} \\
                        NUM:    & \begin{tfs}{sg} \end{tfs} 
                        \end{tfs} \\
        SEM:    & \begin{tfs}{sem} PRED: & \begin{tfs}{love} \end{tfs} \end{tfs}
\end{tfs}
\end{array}
\]
}
Rules:
{\tiny
\begin{eqnarray}
\label{rule:snpvp}
\begin{tfs}{phrase}
        CAT:    & \begin{tfs}{s} \end{tfs} \\
        AGR:    & \tag{1}\\
        SEM:    & \tag{2}\begin{tfs}{sem}
                        ARG1: & \tag{3}
                \end{tfs}
\end{tfs}
&\longrightarrow &
\begin{tfs}{sign}
        CAT:    & \begin{tfs}{n} CASE: & \begin{tfs}{nom}\end{tfs} \end{tfs} \\
        AGR:    & \tag{1} \\
        SEM:    & \begin{tfs}{sem} PRED: & \tag{3} \end{tfs}
\end{tfs}
\begin{tfs}{sign}
        CAT:    & \begin{tfs}{v} \end{tfs} \\
        AGR:    & \tag{1}\\
        SEM:    & \tag{2}
\end{tfs}
\\
\label{rule:vpvnp}
\begin{tfs}{phrase}
        CAT:    & \begin{tfs}{v} \end{tfs} \\
        AGR:    & \tag{1} \\
        SEM:    & \tag{2}\begin{tfs}{sem} ARG2: & \tag{3} \end{tfs}
\end{tfs}
&\longrightarrow &
\begin{tfs}{sign}
        CAT:    & \begin{tfs}{v} \end{tfs} \\
        AGR:    & \tag{1} \\
        SEM:    & \tag{2}
\end{tfs}
\begin{tfs}{sign}
        CAT:    & \begin{tfs}{n} CASE: & \begin{tfs}{acc}\end{tfs} \end{tfs} \\
        SEM:    & \begin{tfs}{sem} PRED: \tag{3} \end{tfs}
\end{tfs}
\end{eqnarray}
}
\caption{An example grammar}
\label{fig:grammar}
\end{figure}

In what follows we define the notion of {\em derivation\/} (or {\em
rewriting}) with respect to TFS-based grammars. Informally, this
relation (denoted `$\derives$') is defined over MRSs such that
$\sigma\derives\rho$ iff $\rho$ can be obtained from $\sigma$ by
successive application of grammar rules. The reader is referred to,
e.g.,~\cite{sikkel,deductive-parsing,shuly:tr-lcl-95-1} for a detailed
formulation of this concept for a variety of formalisms.

To define derivations we first define {\em immediate derivation}.
Informally, two MRSs $A$ and $B$ are related by immediate derivation
if there exists some grammar rule $\rho$ that licenses the
derivation. $\rho$ can license a derivation by some MRS $R$ that it
subsumes; the head of $R$ must be identified with some element $i$ in
$A$, and the body of $R$ must be identified with a sub-structure of
$B$, starting from $i$.  The parts of $A$ prior to and following $i$
remain intact in $B$.  Note that $R$ might carry reentrancies from
$A$ to $B$: if a path $\pi_2$ leaving the $i$-th element of $A$ is
reentrant with some path $\pi_1$ leaving the $a$-th element, and
$\pi_2$, starting from the head of $R$, is reentrant with $\pi_3$ in
some element $b$ in $R$, then $\pi_1$ and $\pi_3$ are reentrant in
$B$, starting from the elements in $B$ that correspond to $a$ and $b$,
respectively.

\begin{definition}
A MRS $A=\langle A_1,\ldots A_k \rangle$ {\bf immediately derives} a
MRS $B=\langle B_1,\ldots B_m \rangle$ (denoted $A \derivesim B$) iff
there exist a rule $\rho \in \rules$ of length $n$ and a MRS $R \subsumed
\rho$, such that:
\begin{itemize}
\item
$m=k+n-2$
\item
$R$'s head is identified with some element $i$ of $A$: $R^1 = A^i$;
\item
$R$'s body is identified with a sub-structure of $B$: $R^{2\ldots n}
= B^{i \ldots i+n-2}$ 
\item
The first $i-1$ elements of $A$ and $B$ are identical: $A^{1\ldots
i-1} = B^{1 \ldots i-1}$;
\item
The last $k-i$ elements of $A$ and $B$ are identical: $A^{i+1
\ldots k} = B^{m-(k-i+1) \ldots m}$.
\end{itemize}
The reflexive transitive closure of `$\derivesim$', denoted
`$\derivess$', is defined as follows: $A \derivess A''$ if
$A = A''$ or if there exists $A'$ such that
$A \derivesim A'$ and $A' \derivess A''$. 
\end{definition} 
\begin{definition}
A MRS $A$ {\bf derives} a MRS $B=$ (denoted $A \derives B$) iff there
exist MRSs $A',B'$ such that $A \subsumes A'$, $B \subsumes B'$ and
$A' \derivess B'$.
\end{definition}

Immediate derivation is based on the more traditional notion of {\em
substituting\/} some symbol which constitutes the head of some rule
with the body of the rule, leaving the context intact. However, as our
rules are based on TFSs, the context of the ``symbol'' to be
substituted might be affected by the substitution. To this end we
require identity, and not only unifiability, of the contexts.  MRSs
related by derivations should be viewed as being ``as specific as
needed'', i.e., containing all the information that is added by the
rule that licenses the derivation.  This is also the reason for the
weaker conditions on the `$\derives$' relation: it allows an MRS $A$
to derive an MRS $B$ if there is a sequence of immediate derivations
that starts with a sepcification of $A$ and ends in a specification of
$B$.

\begin{figure}[hbt]
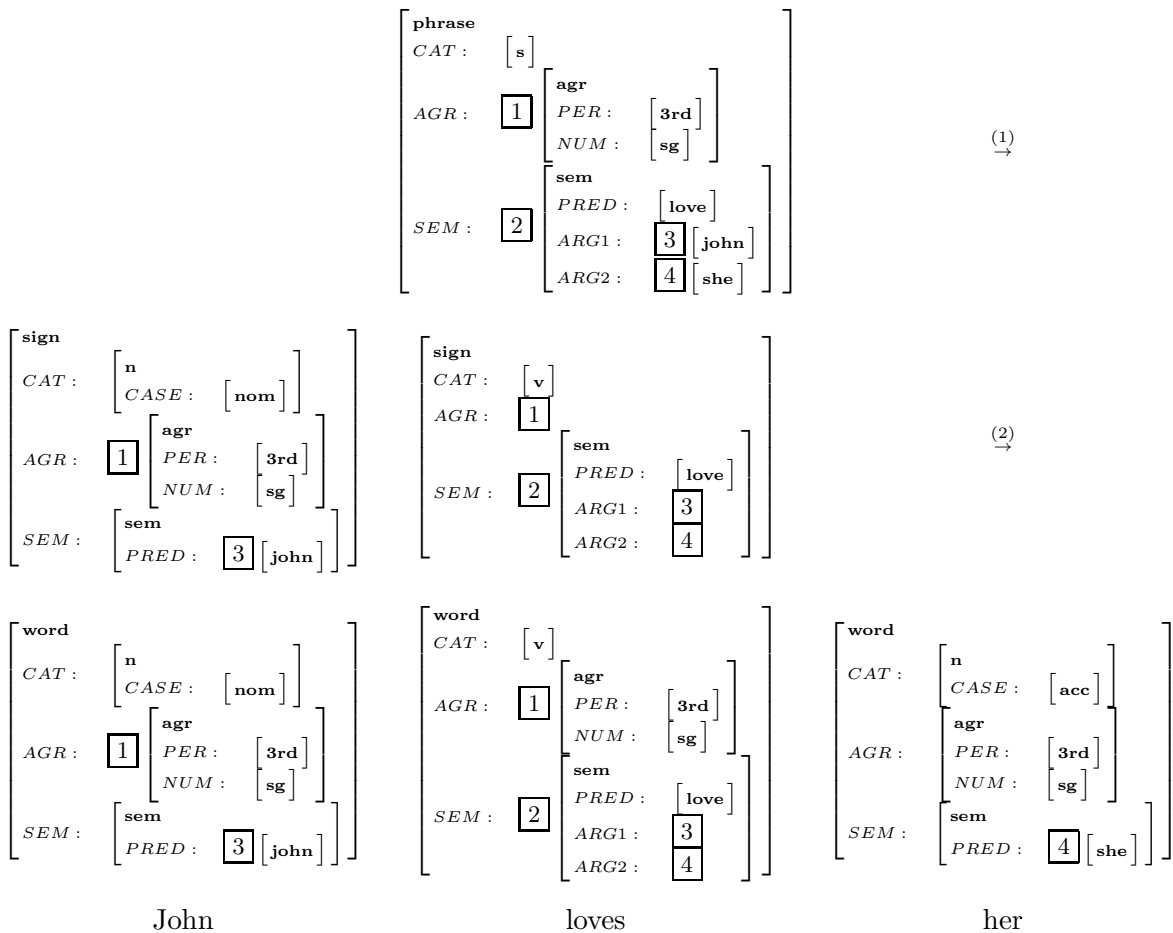

{\tiny
\[
\begin{array}{ccc}
 &
\begin{tfs}{phrase}
        CAT:    & \begin{tfs}{s} \end{tfs} \\
        AGR:    & \tag{1}\begin{tfs}{agr}
                        PER:    & \begin{tfs}{3rd} \end{tfs} \\
                        NUM:    & \begin{tfs}{sg} \end{tfs} 
                        \end{tfs}\\
        SEM:    & \tag{2}\begin{tfs}{sem}
                        PRED: & \begin{tfs}{love}\end{tfs}\\
                        ARG1: & \tag{3}\begin{tfs}{john}\end{tfs}\\
                        ARG2: & \tag{4}\begin{tfs}{she}\end{tfs}
                \end{tfs}
\end{tfs}
& \stackrel{\mbox{(\ref{rule:snpvp})}}{\derivesim}
\\
\\
\begin{tfs}{sign}
        CAT:    & \begin{tfs}{n} CASE: & \begin{tfs}{nom}\end{tfs} \end{tfs} \\
        AGR:    & \tag{1} \begin{tfs}{agr}
                        PER:    & \begin{tfs}{3rd} \end{tfs} \\
                        NUM:    & \begin{tfs}{sg} \end{tfs} 
                        \end{tfs}\\
        SEM:    & \begin{tfs}{sem} 
                        PRED: & \tag{3} \begin{tfs}{john}\end{tfs}
                \end{tfs}
\end{tfs} &
\begin{tfs}{sign}
        CAT:    & \begin{tfs}{v} \end{tfs} \\
        AGR:    & \tag{1}\\
        SEM:    & \tag{2}\begin{tfs}{sem}
                        PRED: & \begin{tfs}{love}\end{tfs}\\
                        ARG1: & \tag{3}\\
                        ARG2: & \tag{4}
                \end{tfs}
\end{tfs}
& \stackrel{\mbox{(\ref{rule:vpvnp})}}{\derivesim}
\\
\\
\begin{tfs}{word}
        CAT:    & \begin{tfs}{n} CASE: & \begin{tfs}{nom} \end{tfs}\end{tfs} \\
        AGR:    & \tag{1}\begin{tfs}{agr}
                        PER:    & \begin{tfs}{3rd} \end{tfs} \\
                        NUM:    & \begin{tfs}{sg} \end{tfs} 
                        \end{tfs} \\
        SEM:    & \begin{tfs}{sem} 
                        PRED: & \tag{3}\begin{tfs}{john}\end{tfs} 
                        \end{tfs}
\end{tfs}
&
\begin{tfs}{word}
        CAT:    & \begin{tfs}{v} \end{tfs} \\
        AGR:    & \tag{1}\begin{tfs}{agr}
                        PER:    & \begin{tfs}{3rd} \end{tfs} \\
                        NUM:    & \begin{tfs}{sg} \end{tfs} 
                        \end{tfs} \\
        SEM:    & \tag{2}\begin{tfs}{sem} 
                        PRED: & \begin{tfs}{love} \end{tfs}\\
                        ARG1: & \tag{3}\\
                        ARG2: & \tag{4} 
                        \end{tfs}
\end{tfs}
&
\begin{tfs}{word}
        CAT:    & \begin{tfs}{n} CASE: & \begin{tfs}{acc} \end{tfs}\end{tfs} \\
        AGR:    & \begin{tfs}{agr}
                        PER:    & \begin{tfs}{3rd} \end{tfs} \\
                        NUM:    & \begin{tfs}{sg} \end{tfs} 
                        \end{tfs} \\
        SEM:    & \begin{tfs}{sem} 
                        PRED: & \tag{4}\begin{tfs}{she} \end{tfs} 
                        \end{tfs}
\end{tfs} 
\\
\\
\mbox{\normalsize John} & \mbox{\normalsize loves} & \mbox{\normalsize her}
\end{array}
\]
}
\caption{A leftmost derivation}
\label{fig:derivation}
\end{figure}

Figure~\ref{fig:derivation} depicts a derivation of the
string ``John loves her'' with respect to the example grammar. The
scope of reentrancy tags should be limited to one MRS, but we use
the same tags across different MRSs to emphasize the flow of
information during derivation.

\begin{definition}[Language]
The {\bf language} of a grammar $G$ is $L(G)=\{w\in\words^*\mid w=w_1
\cdots w_n$ and $A_s \derives PT_w(1,n)\}$.
\end{definition}
The derivation example of figure~\ref{fig:derivation} shows that the
sentence ``John loves her'' is in the language of the example grammar,
since the derivation starts with a TFS that is more specific than the
initial symbol and ends in a specification of the lexical entries of
the sentences' words.

{\em Parsing\/} is a computational process triggered by some input string of
words $w=w_1 \cdots w_n$ of length $n \ge 0$. For the following
discussion we fix a particular grammar $G=(\rules,A_s)$ and a
particular input string $w$ of length $n$.  A {\em state} of the
computation consists of a set of {\em items}.
\begin{definition}[Items]
An {\bf item} is a tuple $[i,\sigma,j,k]$, where $i,j \in \ns$,
$i \le j$, $\sigma$ is an MRS and $0 < k \le |\sigma|$. 
an item is {\bf active} if $k<|\sigma|$, otherwise
it is {\bf complete}. \items\ is the collection of all items.
\end{definition} 
If $[i,\sigma,j,k]$ is an item, $\sigma^{1\ldots k}$ is said to {\em span} the
input from position $i+1$ to position $j$ (the parsing invariant below
motivates this term). $\sigma$ and $k$ can be seen as a representation
of a {\em dotted rule}, or {\em edge}: during parsing all generated
items are such that $\sigma$ is (possibly more specific than) some
grammar rule. $k$ is a position in $\sigma$, indicating the location
of the {\em dot}. The part of $\sigma$ prior to the dot was already
seen; the part following the dot is still expected. When the entire
body of $\sigma$ is seen, the edge becomes complete.

A {\em computation\/} amounts to successively generating items; we
assume that item generation is done through a finite set of
deterministic {\em operations} that create an item on the basis of
previously generated (zero or more) items. Also, if an item was
generated on the basis of some existing items, those items are not
used again by the same operation. This assumption is realized by an
important class of parsing algorithms known as {\em chart parsers}.  A
computation is {\em terminating\/} if and when no new items can be
generated. A computation is {\em successful\/} if, upon termination,
a complete item that spans the entire input and contains the initial symbol
was generated: the final state of the computation should contain the
item $[0,\sigma,n,1]$, where $A \sim A_s$ and $n$ is the input's length.  Different algorithms
assign different meanings to items, and generate them in various
orders (see, e.g.,~\cite{deductive-parsing,sikkel}). To be as general
as possible, we only assume that the following invariant holds:
\paragraph{Parsing invariant} {\em
In a computation triggered by $w$, if an item $[i,\sigma,j,k]$ is
generated then $\sigma^{1\ldots k} \derives PT_w(i+1,j)$.
}
\\
One immediate consequence of the invariant is that for all the items
$[i,\sigma,j,k]$ generated when parsing $w$, $0 \le i \le j  \le |w|$.

A parsing algorithm is required to be {\em correct}:

\paragraph{Correctness} {\em
A computation triggered by $w$ is successful iff $w \in L(G)$.
}
\\
Although~\cite{shieber92} uses a different notation
than~\cite{shuly:tr-lcl-95-1}, this property is proven by both.

\section{Termination of parsing}
\label{application}
It is well-known (see, e.g.,~\cite{parsing-as-deduction,johnson88})
that unification-based grammar formalisms are Turing-equivalent, and
therefore the parsing problem is undecidable in the general case.
However, for grammars that satisfy a certain restriction, termination
of the computation can be guaranteed.  We make use of the well-foundedness
of subsumption (section~\ref{wf}) to prove that parsing is terminating
for {\em off-line parsable} grammars.

To assure efficient computation and avoid maintenance of redundant items, many
parsing algorithms employ a mechanism called {\em subsumption check}
(see, e.g.,~\cite{shieber92,sikkel}) to filter out certain generated
items. Define a (partial) order over items: $[i_1,\sigma_1,j_1,k_1]
\preceq [i_2,\sigma_2,j_2,k_2]$ iff $i_1=i_2,j_1=j_2,k_1=k_2$ and
$\sigma_{1} \sqsubseteq \sigma_{2}$.  The subsumption
filter is realized by preserving an item $x$ only if no item $x'$ such
that $x' \preceq x$ was generated previously. Thus, for all items that
span the same substring, only the most general one is maintained.
\cite{shieber92,shuly:tr-lcl-95-1} prove that by admitting the
subsumption check, the correctness of the
computation is preserved.

{\em Off-line parsability\/} was introduced by~\cite{lfg} and was adopted
by~\cite{parsing-as-deduction}, according to which ``A grammar is
off-line parsable if its context-free skeleton is not infinitely
ambiguous''. As~\cite{johnson88} points out, this restriction (which he
defines in slightly different terms) ``ensures that the number of
constituent structures that have a given string as their yield is
bounded by a computable function of the length of that string''.  The
problem with this definition is demonstrated by~\cite{haas}: ``Not every
natural unification grammar has a context-free backbone''.

A context-free backbone is inherent in LFG, due to the separation of
c-structure from f-structure and the explicit demand that the
c-structure be context-free.  However, this notion is not well-defined
in HPSG, where phrase structure is encoded within feature structures
(indeed, HPSG itself is not well-defined in the formal language
sense).  Such a backbone is certainly missing in Categorial Grammar,
as there might be infinitely many categories. \cite{shieber92}
generalizes the concept of off-line parsability but doesn't prove that
parsing with off-line parsable grammars is terminating. We use an
adaptation of his definition below and provide a proof.

\begin{definition}[Finite-range decreasing functions]
A total function $F:D\rightarrow D$, where $D$ is a partially-ordered
set, is {\bf finite-range decreasing} (FRD) iff the range of $F$
is finite and for every $d\in D,F(d)\preceq d$.
\end{definition}

\begin{definition}[Strong off-line parsability]
A grammar is {\bf strongly off-line parsable} iff there exists an
FRD-function $F$ from MRSs to MRSs (partially ordered by subsumption)
such that for every string $w$ and different MRSs $\sigma,\rho$ such
that $\sigma \derives \rho$, if $\sigma\derives PT_w(i+1,j)$ and
$\rho\derives PT_w(i+1,j)$ then $F(\sigma) \neq F(\rho)$.
\end{definition}
Strong off-line parsability guarantees that any particular sub-string
of the input can only be spanned by a finite number of MRSs:
if a grammar is strongly off-line parsable, there can not exist an
infinite set $S$ of MRSs, such that for some $0 \le i \le j \le |w|$,
$s \derives PT_w(i+1,j)$ for every $s\in S$.  If such a set existed,
$F$ would have mapped its elements to the set $\{F(s) \mid s \in
S\}$. This set is infinite since $S$ is infinite and $F$ doesn't map
two different items to the same image, and thus the finite
range assumption on $F$ is contradicted.

As \cite{shieber92} points out, ``there are non-off-line-parsable
grammars for which termination holds''.  We use below a more general
notion of this restriction: we require that $F$ produce a different
output on $\sigma$ and $\rho$ only if they are incomparable with
respect to subsumption. We thereby extend the class of grammars for
which parsing is guaranteed to terminate (although there still remain
decidable grammars for which even the weaker restriction doesn't hold).

\begin{definition}[Weak off-line parsability]
A grammar $G$ is {\bf weakly off-line parsable} iff there exists an
FRD-function $F$ from MRSs to MRSs (partially ordered by subsumption)
such that for every string $w$ and different MRSs $\sigma,\rho$ such
that $\sigma \derives \rho$, if $\sigma\derives PT_w(i+1,j)$,
$\rho\derives PT_w(i+1,j)$, $\sigma\not\sqsubseteq\rho$ and
$\rho\not\sqsubseteq\sigma$, then $F(\sigma) \neq F(\rho)$.
\end{definition}
Clearly, strong off-line parsability implies weak off-line parsability.
However, as we show below, the inverse implication does not hold.

We now prove that weakly off-line parsable grammars guarantee
termination of parsing in the presence of acyclic MRSs. We prove that
if these conditions hold, only a {\em finite} number of different items can
be generated during a computation. The main idea is the following: if
an infinite number of different items were generated, then an infinite
number of different items must span the same sub-string of the input
(since the input is fixed and finite). By the parsing invariant, this
would mean that an infinite number of MRSs derive the same sub-string
of the input. This, in turn, contradicts the weak off-line parsability
constraint.

\begin{theorem}
\label{olp}
If $G$ is weakly off-line parsable and MRSs are acyclic then every
computation terminates.
\end{theorem}
\proof{}
Fix a computation triggered by $w$ of length $n$.  By the consequence
of the parsing invariant, the
indices that determine the span of items are limited ($0 \le i \le j
\le n$), as are the dot positions ($0 < k \le |\sigma|$).
It remains to show that for every selection
of $i$, $j$ and $k$, only a finite number of MRSs are generated. Let
$x=[i,\sigma,j,k]$ be a generated item. Suppose another item is
generated where only the MRS is different: $x'=[i,\rho,j,k]$ and
$\sigma\neq\rho$. If $\sigma \sqsubseteq \rho$, $x'$ will not be
preserved because of the subsumption test.  If $\rho \sqsubseteq
\sigma$, $x$ can be replaced by $x'$. There is only a finite number
of such replacements, since subsumption is well-founded for acyclic
MRSs. Now suppose $\sigma \not\sqsubseteq \rho$ and $\rho
\not\sqsubseteq \sigma$.  By the parsing invariant, $\sigma^{1\ldots k}
\derives PT_w(i+1,j)$ and $\rho^{1\ldots k} \derives PT_w(i+1,j)$.
Since $G$ is weakly off-line parsable, $F(\sigma^{1\ldots k}) \neq
F(\rho^{1\ldots k})$. Since the range of $F$ is finite, there are only
finitely many items with equal span that are pairwise incomparable.
Since only a finite number of items can be generated and the
computation uses a finite number of operations, every computation ends
within a finite number of steps.
\qed

The above proof relies on the well-foundedness of subsumption, and
indeed termination of parsing is not guaranteed by weak off-line
parsability for grammars based on cyclic TFSs. Obviously, cycles can
occur during unification even if the unificands are acyclic. However,
it is possible (albeit costly, from a practical point of view) to spot
them during parsing. Indeed, many implementations of logic programming
languages, as well as of unification-based grammars (e.g.,
ALE~\cite{ale}) do not check for cycles. If cyclic TFSs are allowed,
the more strict notion of strong off-line parsability is needed. Under
the strong condition the above proof is applicable for the case of
non-well-founded subsumption as well.

To exemplify the difference between strong and weak off-line
parsability, consider a grammar $G$ that contains 
the following single rule:
\[
\begin{tfs}{t}f: & \tag{1}\end{tfs}
\Rightarrow
\tag{1}\begin{tfs}{t}f: & \bot\end{tfs}
\]
and the single lexical entry, $w_1$, whose category is:
\[
Cat(w_1)= \begin{tfs}{t}f: & \bot\end{tfs}
\]
This lexical entry can be derived by an infinite number of TFSs:
\[
\ldots \derivesim
\begin{tfs}{t}f: & \begin{tfs}{t}f: & \begin{tfs}{t}f: & \bot\end{tfs}\end{tfs}\end{tfs} \derivesim
\begin{tfs}{t}f: & \begin{tfs}{t}f: & \bot\end{tfs}\end{tfs} \derivesim
\begin{tfs}{t}f: & \bot\end{tfs} =
Cat(w_1)
\]
%
It is easy to see
that no FRD-function can distinguish (in pairs) among these TFSs, and
hence the grammar is not strongly off-line parsable. 
The grammar
is, however, {\em weakly\/} off-line parsable: since the TFSs that
derive each lexical entry form a subsumption chain, the antecedent of
the implication in the definition for weak off-line parsability never
holds; even trivial functions such as the function that returns the
empty TFS for every input are appropriate FRD-functions. Thus parsing
is guaranteed to terminate with this grammar.

It might be claimed that the example rule is not a part of any grammar
for a natural language. It is unclear whether the distinction between
weak and strong off-line parsability is relevant when ``natural''
grammars are concerned. Still, it is important when the formal,
mathematical and computational properties of grammars are
concerned. We believe that a better understanding of formal properties
leads to a better understanding of ``natural'' grammars as well.
Furthermore, what might be seem un-natural today can be common practice
in the future.

\section*{Acknowledgments}
This work is supported by a grant from the Israeli Ministry of
Science: ``Programming Languages Induced Computational Linguistics''.
The work of the second author was also partially supported by the Fund
for the Promotion of Research in the Technion. We wish to thank the
anonymous referees for their enlightening comments.


\bibliographystyle{fullname}

%
%
\end{document}